\documentclass[runningheads]{llncs}
\usepackage{graphicx}
\usepackage{cite}
\usepackage{amsmath,amssymb,amsfonts}
\usepackage{algorithmic}
\usepackage{graphicx}
\usepackage{float}
\usepackage{textcomp}
\usepackage{seqsplit}
\usepackage{varwidth}
\usepackage{xcolor}
\usepackage{hyperref}
\hypersetup{
    colorlinks=true,
    linkcolor=blue,
    filecolor=blue,      
    urlcolor=blue,
    citecolor=blue
}
\def\BibTeX{{\rm B\kern-.05em{\sc i\kern-.025em b}\kern-.08em
    T\kern-.1667em\lower.7ex\hbox{E}\kern-.125emX}}

\begin{document}
\title{Using Convolutional Neural Networks to Detect Compression Algorithms}
%
%
\author{Shubham Bharadwaj\inst{1}\orcidID{0000-0003-3125-2211}
}

%
%
\institute{Vellore Institute of Technology\\
\email{shubhambharadwaj1108@gmail.com}
}
\maketitle              

\begin{abstract}
Machine learning is penetrating various domains virtually, thereby proliferating excellent results. It has also found an outlet in digital forensics, wherein it is becoming the prime driver of computational efficiency. A prominent feature that exhibits the effectiveness of ML algorithms is feature extraction that can be instrumental in the applications for digital forensics. Convolutional Neural Networks are further used to identify parts of the file. To this end, we observed that the literature does not include sufficient information about the identification of the algorithms used to compress file fragments. With this research, we attempt to address this gap as compression algorithms are beneficial in generating higher entropy comparatively as they make the data more compact. We used a base dataset, compressed every file with various algorithms, and designed a model based on that. The used model was accurately able to identify files compressed using compress, lzip and bzip2.

\end{abstract}

\section{Introduction}

In digital forensics, \textit{data carving} is the act of extracting files directly from some memory media - without any metadata or known filesystem. Conventional techniques use simple heuristics such as magic numbers, headers etc. These techniques do not scale well due to a limited number of supported file types, slow processing speeds as well as an insufficient accuracy\cite{chen2018}.

Recently, machine learning has been applied to the subject, achieving state-of-the-art results both in terms of scale, accuracy and speed. These techniques utilize an efficient feature extraction from files that can be turned into a small image or other representation of the features. The images are then fed to Convolutional Neural Networks (CNNs) to learn to identify parts of files.

These techniques focus on generality to identify files such as documents (.txt, .docx, .ppt, .pdf) and images (.jpg, .png). There is a gap in research when it comes to accurately identifying compressed files and what algorithm was used to compress them. Compression algorithms seek to make data as dense as possible, which will in turn likely yield a higher entropy than a typical file. This in theory could make the identification of the algorithms more difficult than typical image classification problems\cite{PENROSE2013}.

As data carving is a common task in memory forensics, better tooling will most likely help the industry and community with more efficient and accurate detection of compressed files and file chunks.

There is no dataset readily available with a focus on compressed files, therefore an existing dataset, GovDocs\footnote{\href{https://digitalcorpora.org/corpora/files}{https://digitalcorpora.org/corpora/files}}, will be used as a base. Each file will be compressed using various algorithms such as zip, gzip, bzip2, rar etc.

Based off of previous work, a model will be designed and trained.

The tool to create the dataset, train and evaluate the model, as well as using a pre-trained model will be released as part of the paper to allow others to reuse the same dataset and to validate the work.

\subsection{Research Question}

We seek to fill the gap in research when it comes to identifying the compression algorithm used to compress a file fragment. By doing so we hope to answer the following questions.

\begin{itemize}
    \item How do compressed files compare to random data in terms of entropy?
    \item How can a machine-learning system be utilized to detect the algorithm used to compress a file fragment?
\end{itemize}

The hypotheses are:

\textbf{H$_1$}: The machine-learning system achieves an accuracy that is higher than that of random guesses.

\textbf{H$_0$}: The machine-learning system achieves an accuracy equivalent to random guesses.

\subsection{Delimitations}

To limit the scope of the comparison of compressed files and random data, a subset of all available compression algorithms and types of random data will be considered. The analyzed files were selected based on some conditions. These conditions are discussed in greater detail in \ref{method:dataset} and \ref{method:entropy}.

How a machine-learning system can be utilized will be limited to a study of previous state-of-the-art implementations of file fragment identification. Related work is discussed under \ref{introduction:related} and the model implementation under \ref{result:model}.

The accuracy of a random guess is defined to be $12.5\%$ as there are eight classes present in the dataset described in \ref{method:dataset}.

\subsection{Related Work}
\label{introduction:related}
Recently, machine learning and deep learning\cite{2015} has been applied in various problems, from de-noising seismic waves\cite{dutta20193d} to the subject of file fragment identification, achieving state-of-the-art results both in terms of scale, accuracy and speed. These techniques utilize an efficient feature extraction from files that can be turned into a small image or other representation of the features. The images are then fed to convolutional neural networks to learn to identify parts of files.

One article, \textit{File Fragment Classification Using Grayscale Image Conversion and Deep Learning in Digital Forensics}, discusses the application of Convolutional Neural Networks to identify the file type of a file fragment. The presented method achieves accurate results. The produced model was however not tested against compressed files\cite{chen2018}.

Another article, \textit{Using Machine Learning to detect the File Compression or Encryption} focuses on detecting if a file fragment has been compressed or encrypted. It achieves a high accuracy using meta-information such as data received from NIST's Statistical Test Suite. However, it does not mention how to identify the compression algorithm of a file fragment\cite{hahn2018}.

\section{Method}

\subsection{Compression Algorithm Selection}

The compression algorithms and tools to include in the dataset were chosen using the following criteria.

\begin{itemize}
    \item They must be general-purpose
    \item There must be a freely available cross-platform implementation
    \item They must be wide-spread
\end{itemize}

The algorithms and related tools presented in table \ref{fig:method:algorithms} were identified as potential candidates given the above criteria.

\begin{figure}[H]
    \centering
    \begin{tabular}{l|l}
        Algorithm & Tool(s)\\
        \hline
        LZ77 & rar, gzip, zip\cite{HORNEY2013,KONECKI2011} \\
        PPM & 7-zip, rar\cite{HORNEY2013, KONECKI2011} \\
        LZMA & 7-zip\cite{HORNEY2013, KONECKI2011} \\
        BWT & bzip\cite{HORNEY2013, KONECKI2011} \\
        LZW & compress / ncompress\cite{HORNEY2013, KONECKI2011} \\
        lz4 & lz4\footnotemark\\
        brotli & brotli\footnotemark \\
        Deflate & zip, gzip\cite{HORNEY2013} \\
    \end{tabular}
    \caption{Algorithms and Implementations, in no particular order}
    \label{fig:method:algorithms}
\end{figure}

The algorithm implementation differs between tools, such as the implementation of LZ77 for rar, gzip and zip\cite{HORNEY2013}. Therefore, the study will focus on the implementation made available by the respective tools, meaning the tools in figure \ref{fig:method:algorithms2} will be studied 
further.

\begin{figure}[H]
    \centering
    \begin{tabular}{l|l}
        Tool & Version\\
        \hline
        rar & 5.40 \\
        gzip & 1.9 \\
        zip & 3.0 \\
        7-zip & 16.02 \\
        bzip2 & 1.0.6 \\
        ncompress & 4.2.4.5 \\
        lz4 & v1.8.3 \\
        brotli & 1.0.7
    \end{tabular}
    \caption{Implementations chosen for inclusion in the dataset, in no particular order}
    \label{fig:method:algorithms2}
\end{figure}
\subsection{Dataset}
\label{method:dataset}

As a basis for the dataset used in this research lies the publicly available Govdocs1 dataset. It was created with the intention of enabling further research on files and file fragments. The corpora contains one million documents obtained from various web servers on .gov domains.

Since the dataset does not include all the algorithms intended to be studied, a set of tools were developed to deterministically create an adaptation of the dataset using a seed. These tools allow for anyone to download the Govdocs1 dataset and create the dataset used in this research. 

Tools such as zip only support compressing directories whilst tools such as gzip only supports compressing a single file. To make the process more coherent, each file in the dataset is put into a directory. For tools with support for compressing directories, such as rar and zip, this directory is compressed directly. For tools not supporting directories, such as gzip, the tar tool is used to archive the directory. This archive is then compressed.

Each compressed file in the base dataset is split into chunks. As modern file systems store data in chunks or clusters of 4096 bytes\cite{PENROSE2013}, this study will focus on the use of 4096 as the fragment size.


\subsection{Analysis of Entropy}
\label{method:entropy}

Due to the amount of available samples in the dataset and the time required to evaluate a sample, it is impractical to evaluate them all. Instead, a randomly selected sample of 80 chunks (10 per tool) was used, created by using the aforementioned seed of $1.3035772690$

Each chunk in the sample was evaluated using a NIST Statistical Test Suite implementation. NIST's own implementation was not used as it was not deemed suitable for scriptable use due to its interactive interface. It would therefore be impractical to use it for all of the available samples. The results of this paper assumes that the output from the tool is accurate.

NIST has studied the issue of analysing the entropy of random number generators and pseudo-random number generators extensively. They have provided a set of tests that determine whether or not a sequence of bytes can be considered truly random\cite{NIST2010}.

Using this toolset, the entropy of sample files from the dataset was analysed in order to get an idea of whether or not the files may contain any pattern. The procedure is explained in detail below.

The test suite consists of 15 tests, each of which results in a pass or fail. A sample passes a test if it is considered random and fails otherwise. The following tests are performed:

\begin{itemize}
    \item Frequency (Monobit) Test
    \item Frequency Test within a Block
    \item Runs Test
    \item Test for the Longest Run of Ones in a Block
    \item Binary Matrix Rank Test
    \item Discrete Fourier Transform (Specral) Test
    \item Non-Overlapping Template Matching Test
    \item Overlapping Template Matching Test
    \item Maurer's "Universal Statistical" Test
    \item Linear Complexity Test
    \item Serial Test
    \item Approximate Entropy Test
    \item Cumulative Sums (Cumsum) Test
    \item Random Excursions Test
    \item Random Excursions Variant Test
\end{itemize}

By evaluating the entropy of files, one may shed light on available patterns in data which one might exploit for detection. When it comes to compressed and encrypted data, the entropy is usually much higher than that of a non-compressed and non-encrypted file. Previously, researches have argued that this fact makes it difficult to detect features of such high-entropy files\cite{PENROSE2013}.

The produced dataset contains $5653232$ usable 4096 byte sized samples for the algorithms under test. As it was deemed unfeasible to test all samples, sampling was used to select a subset of the dataset to evaluate.

Using the provided tools and the aforementioned seed, an even sample of 80 was created. As the samples are compressed, they should in theory be similar to other samples produced by the same tool, making it meaningless to try many more samples\cite{HORNEY2013}.

The output of the test suite will be compared to two pseudo-random and one random sample. The two pseudo-random samples are taken from the /dev/random file and /dev/urandom file respectively. These samples were created on macOS 11. The random sample was created by random.org\footnote{\href{https://www.random.org/bytes/}{https://www.random.org/bytes/}}. These samples are included in the project's repository.


\section{Results}

\subsection{The Entropy of Compressed Files}

The NIST Statistical Test Suite was used to evaluate two pseudo-random samples and one random sample. These are presented in figure \ref{fig:nist_random}. The pseudo-random sample taken from /dev/urandom had a pass rate of 69\%, the sample from /dev/random 63\% and the truly random sequence passed 69\%.

As for the samples taken from the dataset, the following table, figure \ref{fig:nist_samples} shows the pass rate for each evaluated tool.

\begin{figure}[H]
    \centering
    \begin{tabular}{l|c|c}
        Tool & Pass Rate\\ \hline
        compress & 15\%\\
        lz4 & 30\%\\
        brotli & 42\%\\
        zip & 47\% \\
        gzip & 48\%\\
        rar & 60\%\\
        bzip2 & 63\%\\
        7z & 66\%\\
    \end{tabular}
    \caption{Pass Rate for Evaluated Tools}
    \label{fig:nist_samples}
\end{figure}

\subsection{Model Training and Evaluation}
\label{result:model}

The model is a Convolutional Neural Network based off of previous work by Q. Chen el al\cite{chen2018}. The model takes in a visual representation of a 4096 byte chunk, that is, a 64 by 64 grayscale image. The image is then passed through a series of convolutions and max pools for feature extraction. After feature extraction, the data is passed through two 2048 classification layers ending with a softmax output. It is trained using the Adam optimizer and Sparse Categorical Crossentropy as loss function.\cite{chen2018}. The ReLU\cite{agarap2019deep} activation function was used as a baseline, with hybrid combinations of oscillatory activation functions\cite{noel2021biologically} left as a future work.

\begin{figure*}
    \centering
\includegraphics[scale=0.45]{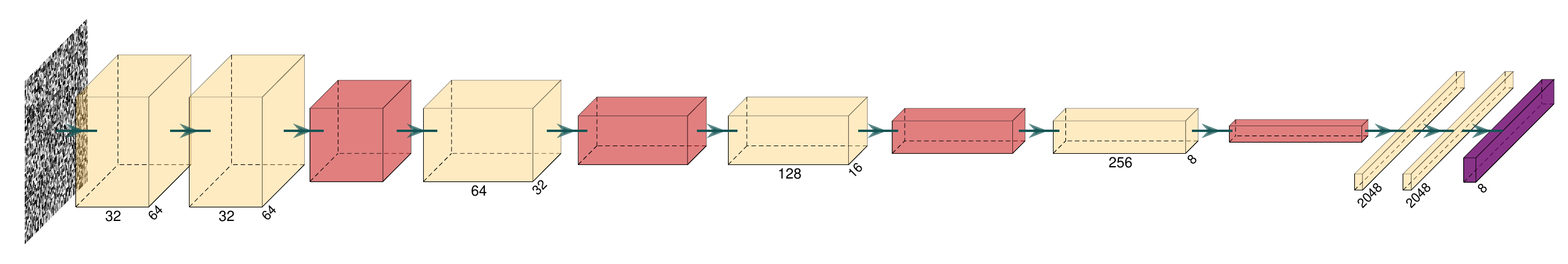}
    \caption{Network Architecture}
    \label{fig:cnn}
\end{figure*}

The model was chosen due to its proven performance in file fragment classification, achieving state-of-the-art\cite{chen2018}. The network architecture is shown in figure \ref{fig:cnn}.

The implementation, related tools and training instructions are available at the paper's repository.

Using the provided tools, all of the available threads of the GovDocs1 dataset were downloaded, compressed and indexed as chunks as mentioned in \ref{method:dataset}. The model was trained on 2 000 000 of these samples (250 000 samples per class) and tested using 200 000 of the samples (25 000 samples per class) (a rough 90\%/10\% split). The used seed for the sampling is $1.3035772690$. The model was set to train for 100 epochs, based on the time required for each epoch and the time we could allocate for this procedure.

\begin{figure}[H]
    \centering
\includegraphics[scale=0.2]{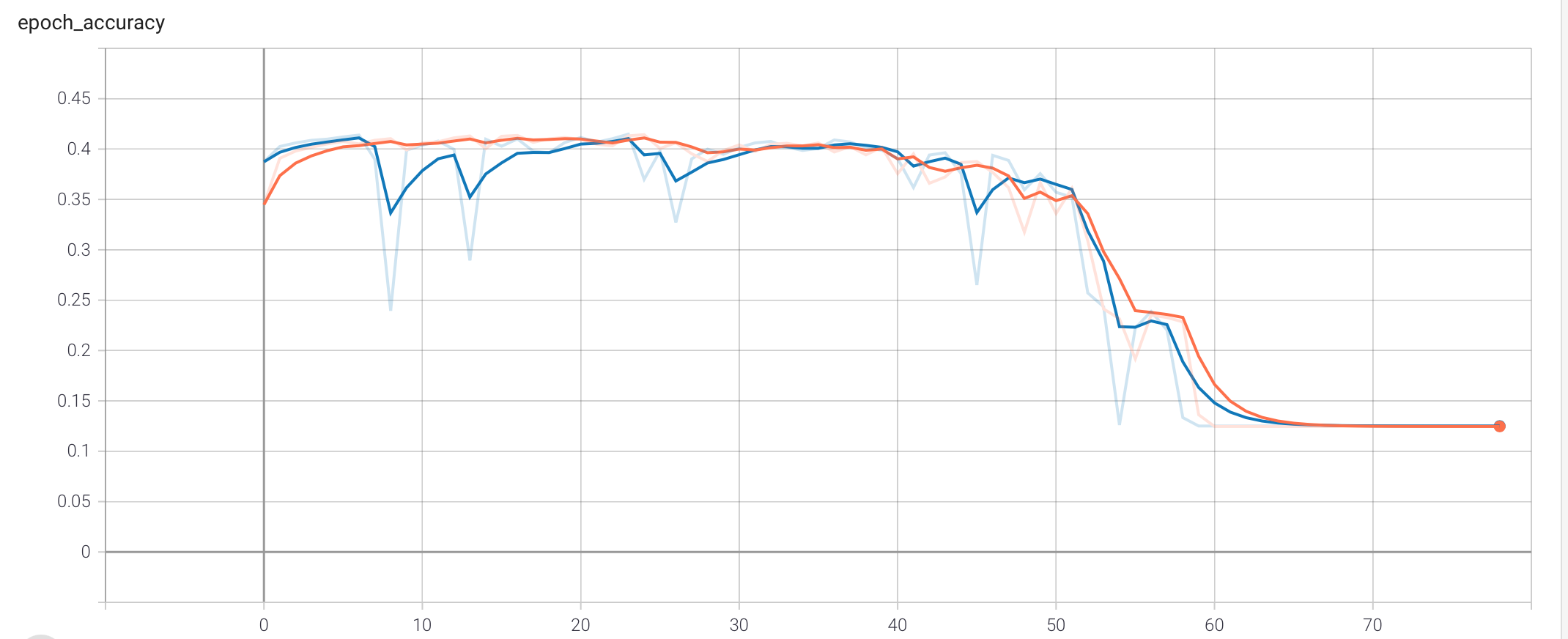}
    \caption{Epoch Accuracy whilst Training}
    \label{fig:epoch-accuracy}
\end{figure}

As can be seen in figure \ref{fig:epoch-accuracy}, the training and validation accuracy never went past 41\%. After 50 epochs, the model started producing worse results. The training was aborted prematurely after 80 epochs as the model did not progress beyond producing an accuracy of 12.43\% for 20 epochs.

The model's state at the fifth epoch (its best performance in terms of validation accuracy) was saved for further analysis.

\begin{figure}
    \centering
\includegraphics[scale=0.25]{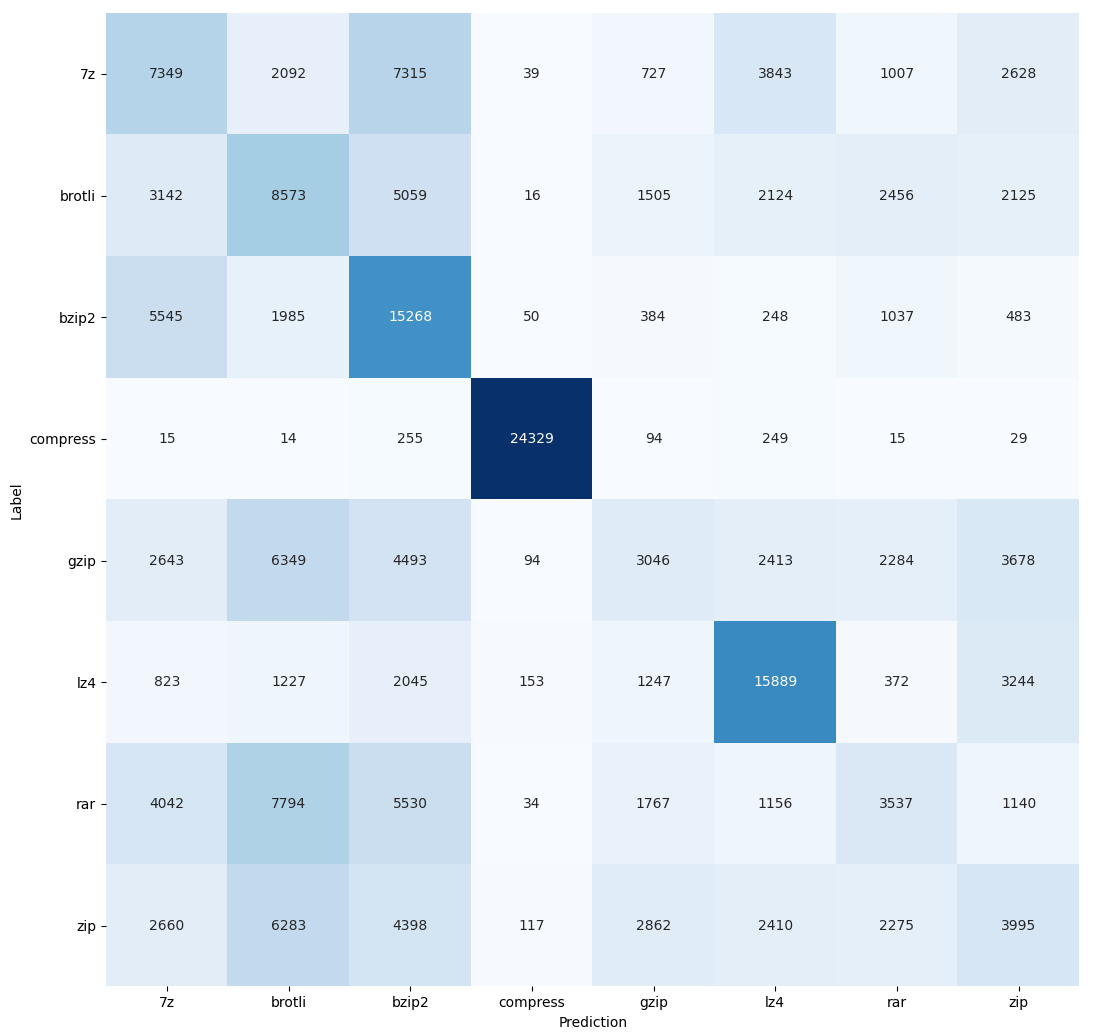}
    \caption{Confusion Matrix}
    \label{fig:confusion-matrix}
\end{figure}

Once trained, the model achieved 41\% accuracy on the validation set. A confusion matrix of the trained model evaluated on 200 000 evenly distributed samples from this set can be seen in figure \ref{fig:confusion-matrix}.

In figure \ref{fig:samples} predicted labels, the prediction confidence and the true label can be seen for the first 25 samples of the validation set. Note that the class ratio in the sample is not representative to the entire dataset.

\begin{figure}[H]
    \centering
    \includegraphics[scale=0.25]{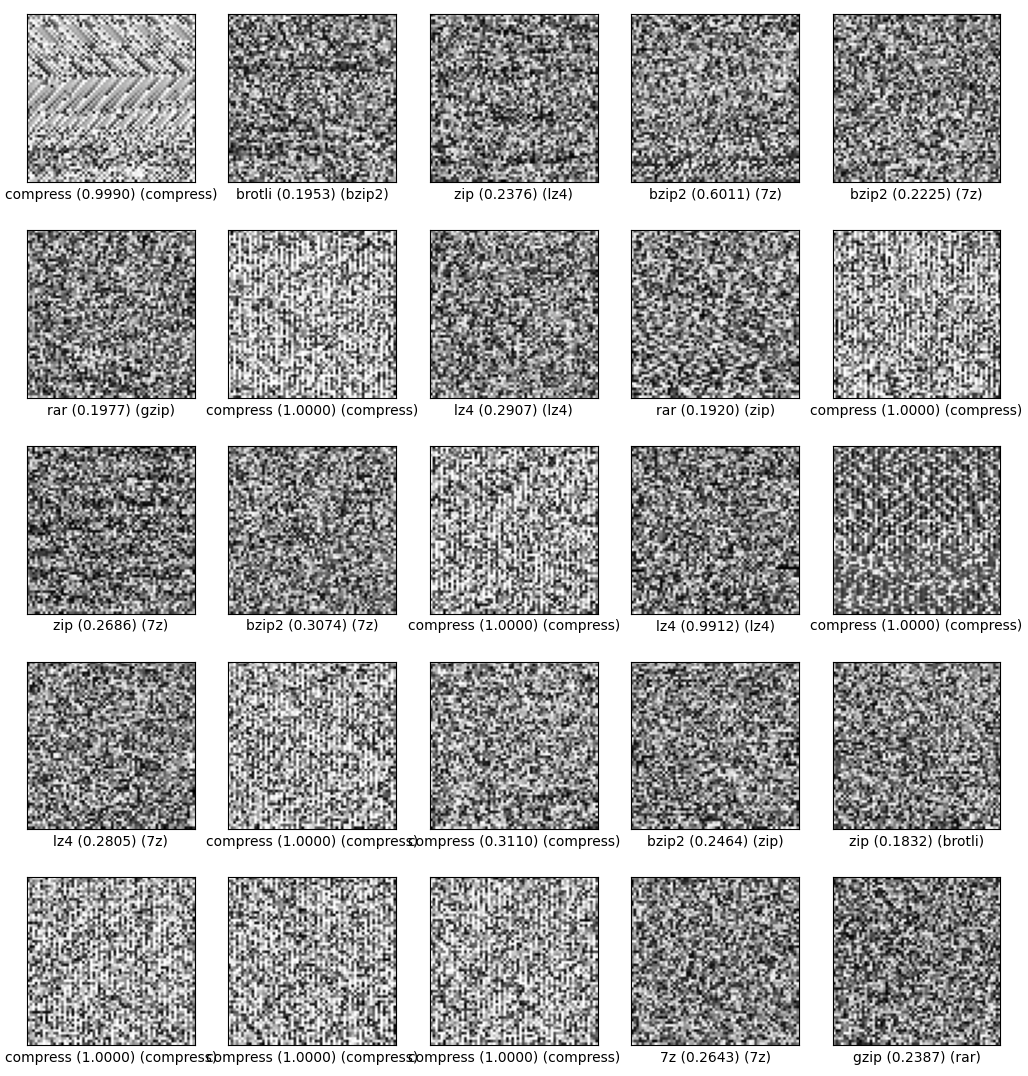}
    \caption{Predicted Samples}
    \label{fig:samples}
\end{figure}

\section{Discussion}

\subsection{The Entropy of Compressed Files}

Given the data presented in figure \ref{fig:nist_random} and figure \ref{fig:nist_samples} one can see that 7z matches the random data distribution of samples from /dev/random. The data from rar and bzip2 are close to random as well. Two of the evaluated tools produce data that is considerably less random than the others - compress and lz4.

In theory, these numbers suggest that a machine learning algorithm should be able to learn the concepts of the structure of the output of compress and lz4, and potentially more tools\cite{lakshmanan2021machine}.

\subsection{Model Training and Evaluation}

The accuracy of the trained model after five epochs was 41\%, thus proving the hypothesis that a machine-learning system can be used to achieve an accuracy higher than that of random guesses.

As presented in figure \ref{fig:confusion-matrix}, the model is able to accurately detect with the compress tool with 97\% accuracy. It is also able to detect bzip2 and lz4 with an accuracy of 61\% and 64\% respectively. The model does confuse 7z and bzip2 about 50\% of the time and also produces somewhat accurate predictions for brotli - the third worst performer in the statistical test.

As can be seen in figure \ref{fig:samples}, these results are not entirely unsurprising given that there are visible patterns in chunks produced by these tools.

This is further confirmed by studying the output of NIST's Statistical Test Suite in figure \ref{fig:nist_samples}. Compress and lz4 both performed the worst of the examined tools. Quite surprisingly, however, the model was able to learn to detect bzip2 even though the samples were hard to distinguish from random.


As mentioned, the training was stopped once the network started producing random guesses. At first, the random guesses may seem to be an error. But as shown previously, several tools produce data virtually indistinguishable from pseudo-random and random data. The model likely learned this fact and started producing equally random guesses. By the end of it's training, it achieved an accuracy of 12.43\%. The model was trained on 8 classes, meaning a random guess would produce an accuracy of 12.5\%.

The performance of the network could be related to its design. By design, the network will downscale the problem and extract features from the file fragment representation\cite{chen2018}. This causes the network to focus on parts of the fragment, instead of considering the entire sample. By utilizing another network such as an LSTM, one could be able to produce more accurate results by letting the network consider the entire file sample and potentially identifying larger patterns\cite{LE2018S118}.

Evaluating the accuracy after training the model using different optimizers and loss functions could have yielded better results.

\subsection{Threats to Validity}
This chapter describes the internal, external and constructional validity and what actions were taken to limit their impacts.

\subsubsection{Internal Validity}
To pre-process, train and evaluate the model, several tools were developed to ease the usage and to increase the reproducability and documentation of the presented findings. These tools may be wrongly developed or build on incorrect assumptions, potentially causing an error in the data. To mitigate this threat, each required step and tool is carefully documented in order to enable others to validate the results. Furthermore, several tests and plots were executed and created to help validate the function of the tools.

During this experiment, time is restraint. Training a large machine learning model requires a lot of time to produce solid results. Due to the time constraint, the trained model used to present the results in this paper might not be accurate or produce accurate figures.

\subsubsection{External Validity}
An implementation of NIST's Statistical Test Suite was chosen due to the ease of use in the context of this dataset. The tool has not been validated by NIST and it might therefore produce incorrect results.

This flaw is mitigated by the fact that we're using relative comparisons on the data produced by the tool.

\subsubsection{Construct Validity}
A potential threat to the construct validity of this work could be the sole use of NIST's Statistical Test Suite. There are more ways of determining the entropy of samples, which could lead to different results than those presented herein.

Another potential threat is the design of the dataset and how the CNN model is trained and evaluated. As we were unable to find any dataset specifically targeting compressed file fragments, a dataset was created. The choices of algorithms, parameters and pre-processing may be incorrect and cause data to be incorrect. To limit the impact of this threat, we have ensured that each compressed file uses the default values wherever possible to produce files as the toolmakers' intended. Furthermore, an established dataset, GovDocs1 was used as the base for the dataset, providing a wide range of potential files for the compression algorithms to work with. This should help limit any potential selection bias.

\subsubsection{Conclusion Validity}
\label{section:conclusion-validity}

Using the presented method, we are confident that one may find use for a machine learning model in the domain of compressed file fragment identification. We do however believe that the time allocated to perform the preparation (designing, preparation of data) and the training and evaluation has the potential to affect the results.

\section{Conclusion}

Compressed file fragments produced by some tools are far from random, whilst some produce data virtually indistinguishable from pseudo-random and random data. The tools compress and lz4 perform far worse than the other evaluated tools in terms of the NIST Statistical Test Suite.

A CNN model may be used and trained on compressed file samples to produce classification with an accuracy of 41\%. The used model was accurately able to identify files compressed using compress, lzip and bzip2.

Future work could focus on developing a new purpose-built CNN, testing hyper-parameters such as oscillatory\cite{noel2021growing} activation functions for feature extraction. The model presented in this paper was initially built for general file fragment classification and was not adapted for compressed file fragment\cite{chen2018}. Another potential improvement would be to train separate models for each class and combine them into a single system - thus potentially increasing the accuracy of some classes. Another extension could involve leveraging LSTMs to capture temporal sequences. Using these networks could enable the model to learn longer distance relationships throughout the files, instead of the simpler patterns detected by CNNs\cite{LE2018S118}. Evaluating the accuracy after training the model using different optimizers and loss functions could  yield superior performance.

%
%
\bibliographystyle{splncs04}
\bibliography{bibliography}

\begin{thebibliography}{10}
\providecommand{\url}[1]{\texttt{#1}}
\providecommand{\urlprefix}{URL }
\providecommand{\doi}[1]{https://doi.org/#1}

\bibitem{agarap2019deep}
Agarap, A.F.: Deep learning using rectified linear units (relu) (2019)

\bibitem{chen2018}
{Chen}, Q., {Liao}, Q., {Jiang}, Z.L., {Fang}, J., {Yiu}, S., {Xi}, G., {Li},
  R., {Yi}, Z., {Wang}, X., {Hui}, L.C.K., {Liu}, D., {Zhang}, E.: File
  fragment classification using grayscale image conversion and deep learning in
  digital forensics. In: 2018 IEEE Security and Privacy Workshops (SPW). pp.
  140--147 (2018). \doi{10.1109/SPW.2018.00029}

\bibitem{dutta20193d}
Dutta, P., Power, B., Halpert, A., Ezequiel, C., Subramanian, A., Chatterjee,
  C., Hari, S., Prindle, K., Vaddina, V., Leach, A., et~al.: 3d conditional
  generative adversarial networks to enable large-scale seismic image
  enhancement. arXiv preprint arXiv:1911.06932  (2019)

\bibitem{hahn2018}
Hahn, D., Apthorpe, N.J., Feamster, N.: Detecting compressed cleartext traffic
  from consumer internet of things devices. CoRR  \textbf{abs/1805.02722}
  (2018), \url{http://arxiv.org/abs/1805.02722}

\bibitem{KONECKI2011}
{Konecki}, M., {Kudelić}, R., {Lovrenčić}, A.: Efficiency of lossless data
  compression. In: 2011 Proceedings of the 34th International Convention MIPRO.
  pp. 810--815 (2011)

\bibitem{lakshmanan2021machine}
Lakshmanan, V.: Machine Learning Design Patterns. O'Reilly Media, Inc, City
  (2021)

\bibitem{LE2018S118}
Le, Q., Boydell, O., {Mac Namee}, B., Scanlon, M.: Deep learning at the shallow
  end: Malware classification for non-domain experts. Digital Investigation
  \textbf{26},  S118 -- S126 (2018).
  \doi{https://doi.org/10.1016/j.diin.2018.04.024},
  \url{http://www.sciencedirect.com/science/article/pii/S1742287618302032}

\bibitem{HORNEY2013}
Mahoney, M.: Data Compression Explained (April 2013),
  \url{http://mattmahoney.net/dc/dce.html}

\bibitem{noel2021growing}
Noel, M.M., Trivedi, A., Dutta, P., et~al.: Growing cosine unit: A novel
  oscillatory activation function that can speedup training and reduce
  parameters in convolutional neural networks. arXiv preprint arXiv:2108.12943
  (2021)

\bibitem{noel2021biologically}
Noel, M.M., Bharadwaj, S., Muthiah-Nakarajan, V., Dutta, P., Amali, G.B.:
  Biologically inspired oscillating activation functions can bridge the
  performance gap between biological and artificial neurons. arXiv preprint
  arXiv:2111.04020  (2021)

\bibitem{PENROSE2013}
Penrose, P., Macfarlane, R., Buchanan, W.J.: Approaches to the classification
  of high entropy file fragments. Digital Investigation  \textbf{10}(4),  372
  -- 384 (2013). \doi{https://doi.org/10.1016/j.diin.2013.08.004},
  \url{http://www.sciencedirect.com/science/article/pii/S174228761300090X}

\bibitem{NIST2010}
Rukhin1, A., Soto2, J., Nechvatal2, J., Smid2, M., Barker2, E., Leigh1, S.,
  Levenson1, M., Vangel1, M., Banks1, D., Heckert1, A., Dray2, J., nd, S.V.: A
  statistical test suite for random and pseudorandom number generators for
  cryptographic applications  (April 2010).
  \doi{https://doi.org/10.6028/NIST.SP.800-22r1a},
  \url{https://nvlpubs.nist.gov/nistpubs/Legacy/SP/nistspecialpublication800-22r1a.pdf}

\bibitem{2015}
Schmidhuber, J.: Deep learning in neural networks: An overview. Neural Networks
   \textbf{61},  85–117 (Jan 2015). \doi{10.1016/j.neunet.2014.09.003},
  \url{http://dx.doi.org/10.1016/j.neunet.2014.09.003}

\end{thebibliography}

\newpage
\onecolumn
\section*{Appendix}

\subsection{}
\begin{figure}[H]
    \centering
    \begin{tabular}{l|c|c|c}
        Test & /dev/urandom & /dev/random & random.org \\ \hline
        Frequency (Monobit) Test & Passed & Passed & Passed \\
        Frequency Test within a Block & Passed & Passed & Passed \\
        Runs Test & Passed & Passed & Passed \\
        Test for the Longest Run of Ones in a Block & Passed & Passed & Passed \\
        Binary Matrix Rank Test & Failed & Failed & Failed \\
        Discrete Fourier Transform (Specral) Test & Passed & Passed & Passed \\
        Non-Overlapping Template Matching Test & Passed & Passed & Passed \\
        Overlapping Template Matching Test & Failed & Failed & Failed \\
        Maurer's "Universal Statistical" Test & Failed & Failed & Failed \\
        Linear Complexity Test & Failed & Failed & Failed \\
        Serial Test & Passed & Passed & Passed \\
        Approximate Entropy Test & Passed & Passed & Passed \\
        Cumulative Sums (Cumsum) Test & Passed & Passed & Passed \\
        Random Excursions Test & Passed & Passed & Passed \\
        Random Excursions Variant Test & Failed & Passed & Passed \\ \hline
        Sum & 10/15 & 11/15 & 11/15
        \end{tabular}
    \caption{Pseudo-Random and Random NIST Results}
    \label{fig:nist_random}
\end{figure}

\subsection{}

\subsection{}
\begin{figure}[H]
    \centering
\includegraphics[scale=0.2]{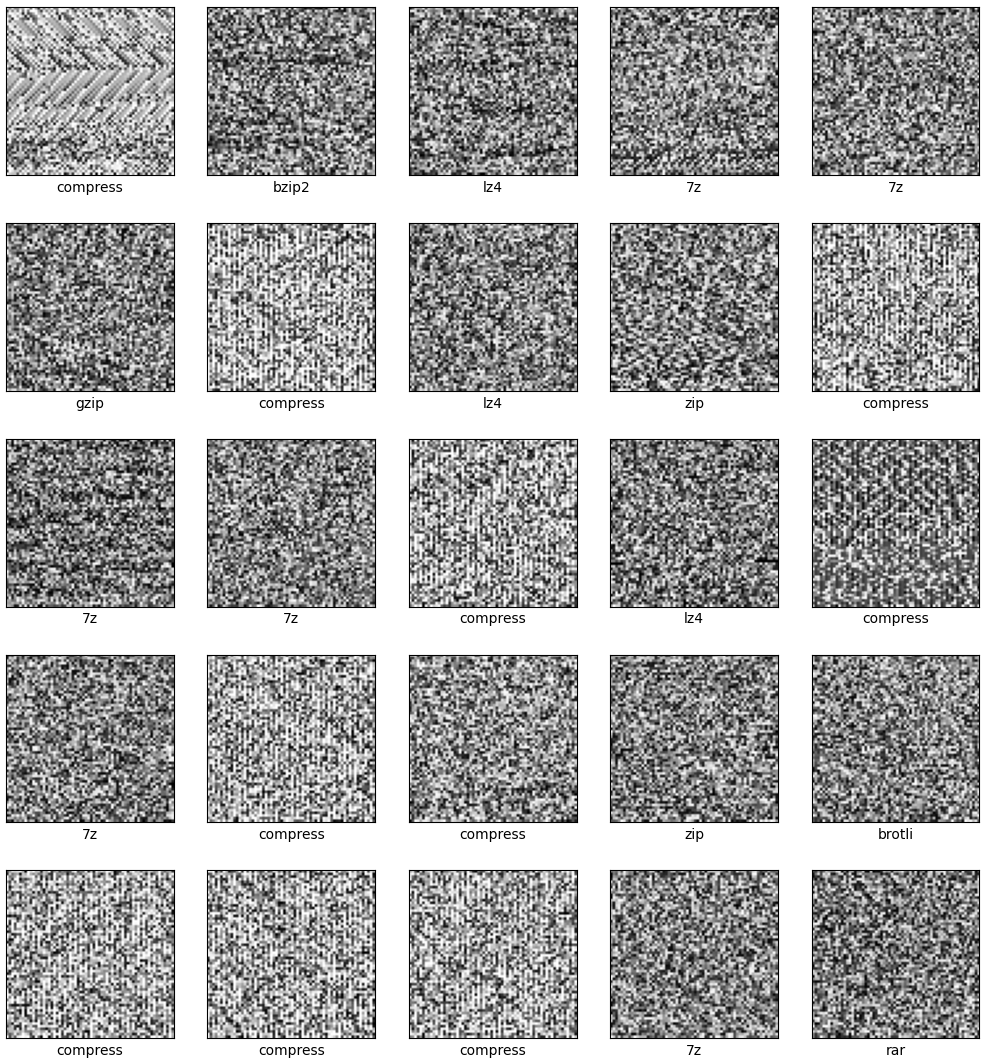}
    \caption{Sample from Dataset}
    \label{fig:samples}
\end{figure}

\end{document}